\documentclass{article}
\usepackage{arxiv}

\usepackage[utf8]{inputenc} 
\usepackage[T1]{fontenc}    
\usepackage{hyperref}       
\usepackage{url}            
\usepackage{booktabs}       
\usepackage{amsfonts}       
\usepackage{nicefrac}       
\usepackage{microtype}      
\usepackage{lipsum}
\usepackage[nobottomtitles]{titlesec}

\usepackage{graphicx}

\usepackage{amssymb,amsmath,amsfonts,amsthm}
\usepackage{bbm}
\usepackage{mathrsfs}
\usepackage{varioref}

\usepackage[ruled,vlined]{algorithm2e}

\usepackage{natbib}
\usepackage[nottoc]{tocbibind}

\usepackage{authblk}

\providecommand{\keywords}[1]{\textbf{\textit{Keywords ---}} #1}

\title{Deep prediction of investor interest: a supervised clustering approach}
\date{}

\author[1, 2]{Baptiste Barreau}
\author[2]{Laurent Carlier}
\author[1]{Damien Challet}
\affil[1]{Chair of Quantitative Finance\\ MICS Laboratory, CentraleSupélec, Université Paris-Saclay, Gif-sur-Yvette, France}
\affil[2]{BNP Paribas Corporate \& Institutional Banking\\
  Global Markets Data \& Artificial Intelligence Lab\\
  Paris, France}

\affil[]{\texttt{\{baptiste.barreau, laurent.carlier\}@bnpparibas.com}}
\affil[]{\texttt{damien.challet@centralesupelec.fr}\}}

\begin{document}
\maketitle

\begin{abstract}

We propose a novel deep learning architecture suitable for the prediction of investor interest for a given asset in a given time frame. This architecture performs both investor clustering and modelling at the same time. We first verify its superior performance on a synthetic scenario inspired by real data and then apply it to two real-world databases, a publicly available dataset about the position of investors in Spanish stock market and proprietary data from BNP Paribas Corporate and Institutional Banking.    
\end{abstract}


\section{Introduction}
\label{intro}

Predicting investor activity is a challenging problem in Finance. The basic problem can be stated as follows: given many thousands of assets and many thousands of investors, predict which investors will be interested in buying/selling which assets in the next (short) time period. What makes this problem difficult is the large heterogeneity of both investors and assets, compounded by the non-stationary nature of markets and investors and the limited time over which predictions are relevant.

Ad-hoc methods are surprisingly efficient at clustering investors according to their trades in a single asset \citep{tumminello2012identification}. In addition, clusters of investors determined for several assets separately have a substantial overlap \citep{baltakys2018multilayer}, which shows that one may be able to cluster investors for more than a few assets at a time. The activity of a given cluster may systematically depend on the previous activity of some clusters, which can then be used to predict the investment flow of investors \citep{challet2018statistically}. Here, we leverage deep learning to train a single neural network on all the investors and all the assets of a given market and give temporal predictions for each investor. 

The heterogeneity of investors translates into a heterogeneity of investment strategies \citep{tumminello2012identification,musciotto2018long}: for the same set of information, e.g., financial and past activity indicators, investors can take totally different actions. Take for instance the case of an asset whose price has just decreased: some investors will buy it because they have positive long-term price increase expectations and thus are happy to be able to buy this asset at a discount; reversely, some other investors will interpret the recent price decrease as indicative of the future trend or risk and refrain from buying it. 

Formally, in our setting, a strategy $f$ is a mapping from current information $x$ to expression of interest to buy and/or sell a given asset, encoded by a categorical variable $y$: $f: x \mapsto y$. We call here $\mathcal{D} = \{ f_{k}: x \mapsto y \}_{k}$ the set of all the investment strategies that an investor may follow. 
Unsupervised clustering methods suggest that the number of different strategies that describe investors' decisions is finite \citep{musciotto2018long}. We therefore expect our dataset to have a finite number $K$ of clusters of investors, each following a given investment strategy $f_{k}$. Consequently, we expect $\mathcal{D}$ to be such that $|\mathcal{D}| = K$, i.e. $\mathcal{D} = \{ f_{k}: x \mapsto y \}_{k=1,\cdots,K}$. Alternatively, $\mathcal{D}$ can be thought of as the set of distinguishable strategies, which may be smaller than the total number of strategies and which may therefore be considered as an effective set of strategies. At any rate, a suitable algorithm to solve our problem therefore needs to be able to infer the set of investment strategies $\mathcal{D}$.

A simple experiment shows how investors differ. We first transform BNP Paribas CIB bonds' \textit{Request for Quotation} (RFQ) database, along with market data and categorical information related to the investors and bonds, into a dataset of custom-made, proprietary features describing the investors' interactions with all the bonds under consideration. This dataset is built so that each row can be mapped to a triplet \textit{(Investor, Financial Product, Date)}. This structure allows us, for a given investor and at a given date, to provide probabilities of interest in buying and/or selling a given financial product in a given timeframe. As the final decision following an RFQ is not always known, we consider the RFQ itself as the signal of interest in a product. Consequently, we consider a given day to be a \textit{positive event} for a given investor and a given financial product when the investor actually signalled his interest in that product in a window of 5 business days around that day. The reason is twofold: first because bonds are by essence illiquid financial products and second because this increases the proportion of positive events.

 At each date, negative events are randomly sampled in the \textit{(Investor, Financial Product)} pairs that were observed as positive events in the past and that are not positive at this date. Using this dataset, we conduct an experiment to illustrate the \textit{non-universality} of investors, i.e. the fact that investors have distinct investment strategies. The methodology of this experiment is reminiscent of the one used in \citet{sirignano2018universal} to study the universality of equity limit order books.

\begin{figure}[!ht]
	\centerline{\includegraphics[scale = 0.5]{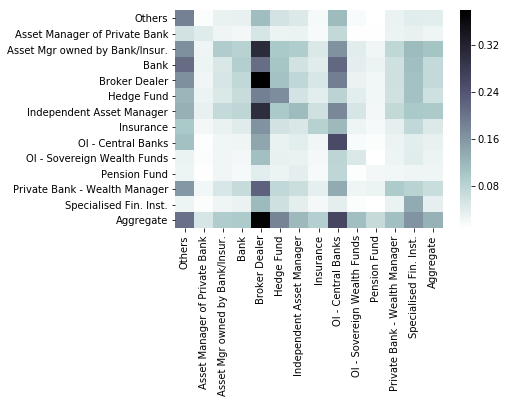}}
	\caption{Universality matrix of investors' strategies: the $y$-axis shows investors' sector used to train a gradient boosting model while the $x$-axis shows investors' sector on which predictions are made using the model indicated on $y$-axis. Scores are average precision, macro-averaged over classes and expressed in percentages.}
	\label{subsector}
\end{figure}

We use a dataset constructed as described above with five months of bonds' RFQ data. We split this dataset into many subsets according to the investors' business sector, e.g. one of these subset contains investors coming from the Insurance sector only.  We consider here only the sectors with a sufficient amount of data samples to train and test a model. The remaining sectors are grouped together under the \textit{Others} flag. Note that this flag is mainly composed of Corporate sectors, such as car industry, media, technology, telecommunications\dots For each sector, some of the latest data is held out, and a gradient boosting model is trained on the remaining data. This model is then used for prediction on the held-out data of the model's underlying sector, and for all the other sectors as well. For comparison purposes, an aggregated model using all sectors at once is also trained and tested in the same way.

Because classes are unbalanced, we compute the average precision score of the obtained results, as advised by \citet{davis2006relationship}, macro-averaged over all the classes, which yields the \textit{universality matrix} shown in Fig. \ref{subsector}. The $y$-axis labels the sector used for training, and the $x$-axis is the section on which the predictions are made.

We observe that some sectors are inherently difficult to predict, even when calibrated on their data only --- this is the case for Asset Managers of Private Banks and Pension Funds. On the contrary, some sectors seem to be relatively easy to predict, e.g. Broker Dealers and, to some extent, Central Banks. Overall, we note that there is always some degree of variability of the scores obtained by a given model ---  no universal model gives good predictions for all the sectors of activity. Thus follows the \textit{non-universality of clients}. In addition, it is worth noting that the aggregated model obtained better performance on some sectors than the models trained on these sectors' data only. As a consequence, a suitable grouping of sectors would improve predictions for some sectors. This observation is in agreement with the above $K$-investment strategies hypothesis.

Following on these hypotheses, this work leverages deep learning both to uncover the structure of similarity between investors, namely the $K$ clusters, or strategies, and to make relevant predictions using each inferred clusters. The advantage of deep learning lies in the fact that it allows to solve both of these tasks at once, and thereby unveils the structure of investors that most closely corresponds to their trading behaviour in a self-consistent way. 

\section{Related work}

This work finds its roots in mixture-of-experts research, which began with \citet{jacobs1991adaptive}, from which we keep the basic elements which drive the structure presented in Section \ref{exnet}, and more particularly the gating and expert blocks. A rather exhaustive history of the research performed on this subject can be found in \citet{yuksel2012twenty}.

The main inspiration for our work is \citet{shazeer2017outrageously}, which, although falling within the \textit{conditional computation} framework, presented the first adaptation of mixture of experts for deep learning models. We build on this work to come up with a novel structure designed to solve the particular problem presented in Section \ref{intro}. As far as we know, the approach we propose is new.  We  use an additional loss term to improve learning of the strategies, reminiscent of the one introduced in \citet{liu1999simultaneous}.

\section{Experts Network}
\label{exnet}

We introduce here a new algorithm, inspired by \citet{shazeer2017outrageously}, which we call the \textit{Experts Network} (ExNet). The ExNet is purposely designed to be able to capture the hypotheses formulated in Section \ref{intro}, i.e. to capture a finite, unknown number $K$ of distinct investment strategies $\mathcal{D} = \{ f_{k}: x \mapsto y \}_{k=1,\cdots,K}$. 

\subsection{Architecture of the network}

\begin{figure}
	\centerline{\includegraphics[scale = 0.5]{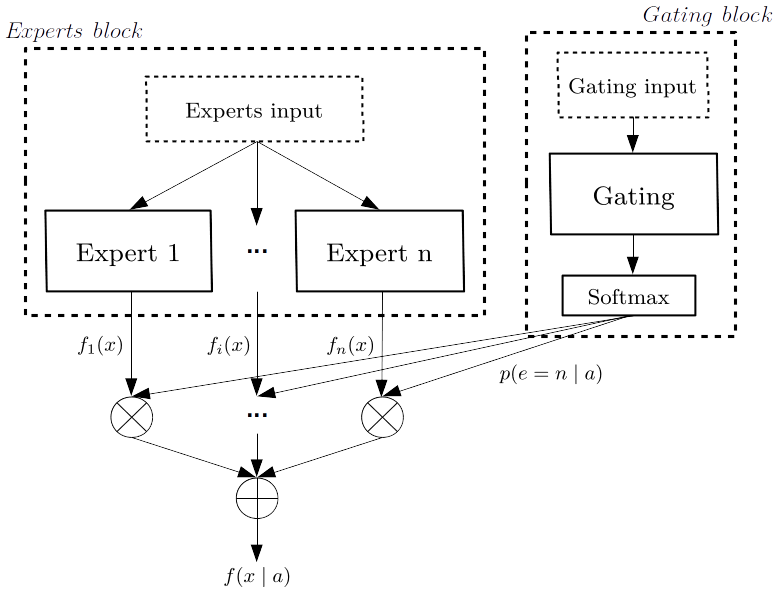}}
	\caption{Global architecture of an ExNet}
	\label{archi}
\end{figure}

The structure of an ExNet, illustrated in Fig. \ref{archi}, comprises two main parts: a \textbf{gating block} and an \textbf{experts block}. Their purposes are the following:
\begin{itemize}
\item The \textbf{gating block} is an independent neural network whose role is to learn how to dispatch investors to $n$ \textit{experts} defined below. This block receives a distinct, categorical input, the \textit{gating input}, corresponding to an encoding of the investors and such that the $i$-th row of the gating input corresponds to the investor indexing the $i$-th row of the experts input. Its output consists in a vector of size $n$ which contains the probabilities that the input should be allocated to the $n$ experts, computed by a softmax activation. 
\item The \textbf{experts block} is made of $n$ independent sub-networks, called \textit{experts}. Each expert receives as input the same data, the \textit{experts input}, corresponding to the features used to solve the classification or regression task at hand, e.g. in our case the features outlined in Section \ref{intro} --- for a given row, the intensity of the investor's interest in the financial asset considered, the total number of RFQ done by the investor, the price and the volatility of the asset\dots As investors are dispatched to the experts through the gating block, each expert will learn a mapping $f: x \mapsto y$ that most closely corresponds to the actions of its attributed investors. The role of an expert is therefore to retrieve a given $f_{k}$, corresponding to one of the $K$ underlying clusters of investors which we hypothesized.
\end{itemize}
The outputs of these two blocks are combined through $f(x|a) = \sum_{i=1}^{n} p(i|a) f_{i}(x)$, where $a$ denotes the investor related to data sample $x$ and $p(i|a)$ is the probability that investor $a$ is assigned to expert $i$. Our goal is that $K$ experts learn to specialize to $K$ clusters. As $K$ is unknown, retrieving all clusters requires that $n \geq K$, i.e. $n$  should be 'large enough'. We will show below that the network ability to retrieve the $K$ clusters is not impacted by high values of $n$; using large $n$ values therefore ensures that the $n \geq K$ condition is respected and only impacts computational efficiency. The described architecture corresponds in fact to a \textit{meta-}architecture. The architecture of the experts is still to be chosen, and indeed any kind of neural network could be used. For the sake of simplicity and computational ease, we use here rather small feed-forward neural networks for the experts, all with the same architecture, but one could easily use experts of different architectures to represent a more heterogeneous space of strategies.

Both blocks are trained simultaneously using gradient descent and backpropagation, with a loss corresponding to the task at hand, be it a regression or classification task, and computed using the final output of the network only, $f(x|a)$. One of the most important features of this network lies in the fact that the two blocks do not receive the same input data. We saw previously that the gating block receives as input an encoding of the investors. As this input is not time-dependent, the gating block of the network can be used \textit{a posteriori} to analyse how investors are dispatched to experts with a single pass of all investors' encodings through this block alone, thereby unveiling the underlying structure of investors interacting in the considered market.

For a given investor $a$, the gating block computes attribution probabilities of investor $a$ to each expert
\begin{center}
$p(x|a) = Softmax\left(W_{experts} * x\right)$,
\end{center}
where $x$ is a trainable $d$-dimensional embedding of the investor $a$, $W_{experts}$ is a trainable $n \times d$-dimensional matrix where the $i$-th row corresponds to the embedding of the corresponding expert, and we define $Softmax(x)_{k} = e^{x_{k}} / \sum_{i} e^{x_{i}}$.

\subsection{Disambiguation of investors' experts mapping}

The ExNet architecture is similar to an ensemble of independent neural networks, where the weighted average is given by the gating block of the network. We empirically noticed that ExNets may assign equal weights to all experts for all investors without additional penalization. To avoid this behaviour, and thereby to help each investor follow a single expert, we introduce an additional loss term
\begin{center}
$L_{\textrm{entropy}} = - \dfrac{1}{|\mathcal{B}|}\sum_{i \in \mathcal{B}} \sum_{j=1}^{n} p^{j}(x|a_{i})\log p^{j}(x|a_{i})$,	
\end{center}
where $\mathcal{B}$ is the current batch of data considered, $n$ is the number of experts, and $p^{j}(x|a_{i})$ is the attribution of investor $a_{i}$ to the $j$-th expert. This loss term corresponds exactly to the entropy of the probability distribution over experts of a given investor. Minimising this loss term will therefore encourage distributions over experts to peak on one expert only. 

\subsection{Helping experts specialize}

Without a suitable additional loss term, the network has a tendency to let a few experts learn the same investment strategy, which also leads to more ambiguous mapping from investors to experts. Thus, to help the network finding different investment strategies and to increase its discrimination power regarding investors, we add a \textit{specialization loss} term, which involves cross-experts correlations, weighted accordingly to their relative attribution probabilities. It is written as: 
\begin{align*}
L_{\textrm{spec}} &= \sum_{i=1}^{n}\sum_{j=1, j \neq i}^{n} w_{i,j} \overline{\rho}_{i,j}\\
\text{with } w_{i, j} &= \dfrac{\overline{p}_{i}\overline{p}_{j}}{\sum_{i=1}^{n}\sum_{j=1, j \neq i}^{n} \overline{p}_{i}\overline{p}_{j}} \text{ if } i \neq j, 0 \text{ else,} \\
\text{and } \overline{p}_{i} &= \dfrac{\sum_{a \in A} p^{a}_{i}}{\sum_{a \in A} \mathbbm{1}_{p^{a}_{i} \neq 0}}.
\end{align*}
Here, $i, j \in \{1,\cdots,n\}$, $\overline{\rho}_{i, j}$ is the batch-wise correlation between experts $i$ and $j$ outputs, averaged over the output dimension, and $\overline{p}_{i}$ is the batch-wise mean attribution probability to expert $i$, with $p^{a}_{i}$ the attribution probability of investor $a$ to expert $i$, computed on the current batch of investors only. The intuition behind this weight is that we want to avoid correlation between experts that were confidently selected by investors, i.e. to make sure that the experts that matter do not replicate the same investment strategy. As the size of the investors clustering around a given expert should not matter in this weighing, we only account for the nonnegative probabilities for all the considered investors in these weights. In some experiments, it was found useful to rescale $L_{\textrm{spec}}$ from $[-1; 1]$ to $[0; 1]$.

This additional loss term is reminiscent of \citet{liu1999simultaneous}. As a matter of fact, in ensembles of machine learning models, negatively correlated models are expected to perform better than positively correlated ones. This can also be expected from the experts of an ExNet, as negatively correlated experts better span the space of investment strategies. As the number of very distinct strategies grow, we can expect to find strategies that more closely match the ones the investors use in the considered market, or the basis functions on which investment strategies can be decomposed.  

\subsection{Uncovering structure from gating}
\label{gating}

Up to this point, we only discuss gating input related to investors. However, as seen above, being able to retrieve the structure of attribution of inputs to experts only requires to use categorical data as input to the gating part of the network after the training phase. We can therefore perform gating on whatever is thought to be suitable --- for instance, it is reasonable to think that bonds investors have different investment strategies depending on the bonds' grades, or depending on the sector of activity of the bonds' issuers. Higher-level details about investors could also be considered, for instance because investment strategies may depend on factors such as the sector of activity of the investor, i.e. whether it is a hedge fund, a central bank or an asset manager, or the region of the investor. The investor dimension could even be totally forgotten, and the gating performed on asset related categories only.

Gating allows one to retrieve the underlying structure of interactions of a given category, or set of categories. One can therefore purposely set categories to study how they relate in the problem one wants to study. This may however impact performance of the model, as chosen categories do not necessarily have distinct decision rules. 

Note also that the initialization of weights in the gating network has a major impact on the future performance of the algorithm. To find relevant clusters, i.e. clusters that are composed of unequivocally attributed categories and that correspond to the original clusters expected in the dataset, categories need to be able to explore many different clusters' configurations before the exploitation of one relevant configuration. To allow for this exploration, the gating block must be initialized so that all the expert weights are fairly evenly initially distributed. In our implementation, we therefore use a random normal initialization scheme for the $d$-dimensional embeddings of the categories and of the experts.

\subsection{Limitations of the approach}

Our approach allows us to treat well a known, fixed base of investors. However, it cannot easily deal with new investors, or, at a higher level, new categories as seen in Section \ref{gating}, as embeddings for these new types of element would need to be trained from scratch. To cope with such situations, we therefore recommend to use sets of fixed categories to describe the evolving ones. For instance, instead of performing gating on investors directly, one can use investors' categories such as sector, region,\dots, that are already present in the dataset and on which we can train embeddings. Doing so improves the robustness of our approach to unseen categories. Note that this is reminiscent of one of the classic problems of recommender systems, known in the literature as the \textit{cold start} problem.

\section{Experiments}

Before testing the ExNet architecture on real data, we first check its ability to recover a known strategy set, to attribute correctly traders to strategies, and finally to classify the interest of traders on synthetic data. We then show how our methodology compares with other algorithms on two different datasets: a dataset open-sourced\footnote{Data is available at \href{url}{https://zenodo.org/record/2573031}} as part of the experiments presented in \citet{gutierrez2019mapping}, and a BNP Paribas CIB dataset. Source code for the experiments on synthetic data and the open-source dataset is provided and can be found at \href{url}{https://github.com/BptBrr/deep\_prediction}.

\subsection{Synthetic data}
\label{simul_data}

\subsubsection{Generating the dataset}

Taking a cue from BNP Paribas CIB bonds' RFQ database, we define three clusters of investors, each having a distinct investment strategy,  which we label as 'high-activity', 'low-activity' and 'medium-activity'. Each cluster contains a different proportion of investors, and each trader within a cluster has the same activity frequency:  the 'high-activity' cluster accounts for roughly $70\%$ of the dataset samples, while containing roughly $10\%$ of the total number of investors. The 'low-activity' cluster accounts for roughly $10\%$ of the samples, while containing roughly $50\%$ of the total number of investors. The 'medium-activity' cluster accounts for the remaining number of samples and investors. In all the clusters, we assume that investors are equally active. 

We model the state of investors as a binary classification task, with a set of $p$ features, denoted by $X\in \mathbb{R}^p$, and a binary output $Y$ representing the fact that a client is interested or not in the considered asset. 
Investor $a$ belonging to cluster $c$ follows the decision rule given by $Y_{a} = (\sigma (w_{a}^{T}X) > U)\in\{0,1\}$, where $w_{a} = w_{c} + b_{a}\in\mathbb{R}^p$, $w_{c}\in\mathbb{R}^p$ being the cluster weights and $b_{a}\sim \mathcal{N}(0, \alpha)\in\mathbb{R}$ an investor-specific bias,  $X_{i} \sim \mathcal{N}(0, 1)$ for $i=1,\cdots,p$, $U$ is distributed according to the uniform distribution on $[0,1]$, and $\sigma$ is the logistic function.

The experiment is conducted using a generated dataset of $100,000$ samples, $500$ investors and $p=5$ features. This dataset is split into train/validation/test sets, corresponding to $70/20/10 \%$ of the whole dataset. 
$\alpha$ is set to $0.5$, and the cluster weights are taken as follows:
\begin{itemize}
\item High-activity cluster: $w_{high} = (5, 5, 0, 0, -5)$
\item Low-activity cluster: $w_{low} = (-5, 0, 5, 0, 5)$
\item Medium-activity cluster: $w_{medium} = (-5, 0, 5, 5, 0)$
\end{itemize}

These weights are chosen so that the correlation between the low- and medium-activity clusters is positive, but both are negatively correlated with the the high-activity cluster. In this way, we build a structure of clusters, core decision rules and correlation patterns that is sufficiently challenging to demonstrate the usefulness of our approach.

\subsubsection{Results}

We examine performance of our proposed algorithm, ExNet, against a benchmark algorithm, LightGBM \citep{ke2017lightgbm}. LightGBM is a popular implementation of gradient boosting, as shown for example by the percentage of top Kaggle submissions that use it. This algorithm is fed with both the experts input of the ExNet and an encoding of the considered investors, used as a categorical feature in the LightGBM algorithm. For comparison purposes, experiments are also performed on a LightGBM model fed with experts input and an encoding of the investors' underlying clusters, i.e. whether the investor belongs to the high-, low- or medium-activity cluster, called \textit{LGBM-Cluster}. 

ExNets are trained using the \textit{cross-entropy} loss, since the problem we want to solve is a classification one. The network is optimized using Nadam \citep{dozat2016incorporating}, a variation of the Adam optimizer \citep{kingma2014adam} using Nesterov's Accelerated Gradient \citep{nesterov1983method}, reintroduced in the deep learning framework by \citet{sutskever2013importance}, and Lookahead \citep{zhang2019lookahead}. For comparison purposes, experiments are also performed on a multi-layer perceptron model fed with the experts inputs concatenated with a trainable embedding of the investors, called \textit{Embed-MLP} --- this model therefore differs from a one-expert ExNet in that this ExNet does not use an embedding of the investor to perform its predictions. All neural network models presented here used the rectified linear unit, $\textrm{ReLU}(x)=\max(x, 0)$, as activation function \citep{nair2010rectified}.

LightGBM, ExNet and Embed-MLP results are shown in Table \ref{exp_results}. They were obtained using a combination of random search \citep{bergstra2012random} and manual fine-tuning. LightGBM-Cluster results used the hyperparameters found for LightGBM. These results correspond to the model which achieved best validation accuracy over all our tests. The LightGBM and LightGBM-Cluster shown had $64$ leaves, a minimum of $32$ samples per leaf, a maximum depth of $10$, a learning rate of $0.005$ and a subsample ratio of $25\%$ with a frequency of $2$. ExNet-Opt, the ExNet which achieved the best validation accuracy, used $16$ experts with three hidden layers of sizes $64$, $128$ and $64$, a dropout rate \citep{srivastava2014dropout} of $40\%$, loss weights $\lambda_{spec}=7e^{-3}$ and $\lambda_{entropy}=1e^{-3}$, a batch size of $1024$, and a learning rate of $7e-3$. The Embed-MLP model shown used two hidden layers of size $128$ and $64$,  a dropout rate of $15\%$, an embedding size $d=64$, a batch size of $64$, and a learning rate of $4.2e^{-5}$.

To study the influence of the number of experts on the performance of ExNets,  we call \textit{ExNet-n} an ExNet algorithm with $n$ experts and vary $n$. These ExNets used experts with no hidden layers, batch-normalized inputs \citep{ioffe2015batch}, and an investor embedding of size $d=64$. These neural networks were trained for $200$ epochs, using early stopping with a patience of $20$. All these experiments were carried out with a learning rate equal to $e^{-3}$ and a batch size of $64$, which was found to lead to satisfactory solutions in all the tested configurations. In other words, we only vary $n$ so as to be able to disentangle the influence of $n$ for an overall reasonably good choice of other hyperparameters. Only the weights attributed to the specialization and entropy losses, $\lambda_{\textrm{spec}}$ and $\lambda_{\textrm{entropy}}$, were allowed to change across experiments. 
\begin{center}
\begin{table}[!ht]
	\begin{tabular}{c c c c c c c}
		\hline
		Algorithm & Train Acc. & Val Acc. & Test Acc. & High Acc. & Medium Acc. & Low Acc. \\
		\hline
		\textbf{LGBM} & 96.38 & 92.05 & 92.41 & 92.85 & 90.47 & 92.34 \\
		\textbf{LGBM-Cluster} & 93.89 & 92.33 & 92.94 & 93.03 & 92.54 & 93.89 \\
		\textbf{Embed-MLP} & 93.87 & 92.88 & 93.19 & 93.20 & 93.14 & 92.24 \\
		\textbf{ExNet-Opt} & \textbf{93.57} & \textbf{92.99} & \textbf{93.47} & \textbf{93.56} & \textbf{93.09} & \textbf{93.17} \\
		\textbf{ExNet-1} & 74.86 & 74.56 & 74.56 & 80.39 & 48.67 & 38.72 \\
		\textbf{ExNet-2} & 90.73 & 90.59 & 90.86 & 91.66 & 87.32 & 82.71 \\
		\textbf{ExNet-3} & 92.73 & 92.50 & 93.06 & 92.97 & 93.47 & 93.89 \\
		\textbf{ExNet-10} & 92.91 & 92.66 & 93.16 & 93.12 & 93.36 & 93.89 \\
		\textbf{ExNet-100} & 92.71 & 92.55 & 93.04 & 92.96 & 93.41 & 93.89 \\		
		\textbf{Perfect model} & 93.62 &  93.51 & 93.71 & 93.75 & 93.52 & 94.82 \\		\hline
	\end{tabular}
	\caption{Experimental results on synthetic data: accuracy of predictions, expressed in percentage, on train, validation and test sets, and on subsets of the test set corresponding to the original clusters to be retrieved.}
	\label{exp_results}
\end{table}
\end{center}
Table \ref{exp_results} contains the results for all the tested implementations. As the binary classification considered here is balanced, we use the accuracy as evaluation metric. This table reports results on train, validation and test splits of the dataset, and a view of the test results on the three different clusters generated. As the generation process provides us with the probabilities of positive events, it is also possible to compute metrics for a model that would output these probabilities, denoted here as \textit{perfect model}, which sets the mark of what good predictive performance is in this experiment. 

We see here that the LGBM implementation fails to completely retrieve the different clusters. LGBM focused on the high-activity cluster and mixed the two others, leading to poorer predictions for both of these clusters and here particularly for the medium-activity one. In comparison, LGBM-Cluster performed significantly better on the medium- and low-activity clusters. Embed-MLP better captured the structure of the problem, but appears to mix the medium- and low-activity clusters as well, albeit getting better predictive performance. ExNet-Opt, found with random search, captured well all clusters and obtained the best overall performances.

Moreover, the ExNet-$n$ experiment shows how the algorithm behaves as $n$ increases. ExNet-$1$ successfully captured the largest cluster in terms of samples, i.e. the high-activity one, partly ignoring the two others, and therefore obtained poor overall performance. ExNet-$2$ behaved as the LGBM experiment, retrieving the high-activity cluster and mixing the remaining two. ExNet-$3$ perfectly retrieved the three clusters, as expected. Even better, the same holds for ExNet-$10$ and ExNet-$100$: this is because the ExNet algorithm, thanks to the additional specialization loss, is not sensitive to the number of experts  even if $n\gg K$, as long as there are enough of them. Thus, when $n\ge K$, the ExNet is able to retrieve the $K$ initial clusters and to predict the interests of these clusters satisfactorily. 

\subsubsection{Further analysis of specialization}

\begin{figure}[!ht]
	\centerline{\includegraphics[scale = 0.5]{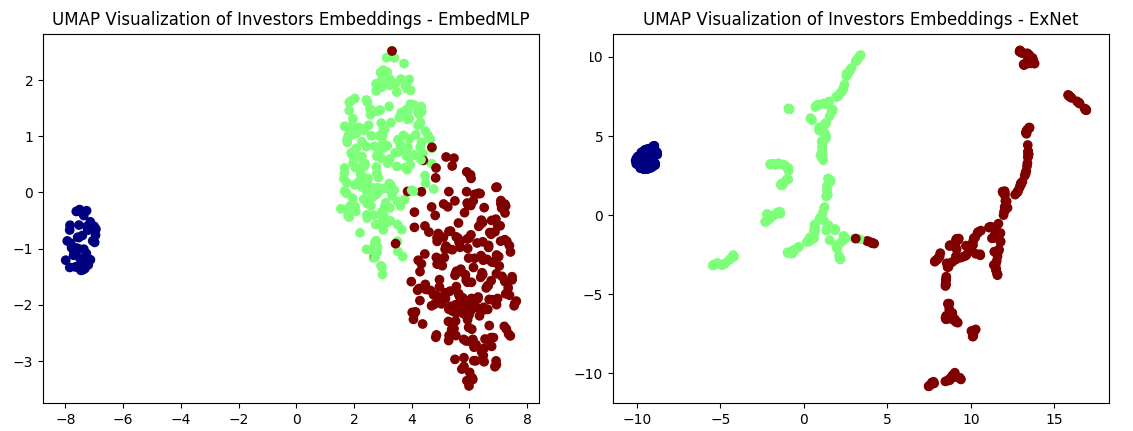}}
	\caption{UMAP visualization of investors embeddings for both Embed-MLP and ExNet algorithms. Colors on the right plot correspond to investors' original clusters: high-activity is shown in blue, medium-activity in green and low-activity in red.}
	\label{mlp_vs_exnet}
\end{figure}

The previous results show that as long as $n \geq K$, the ExNet algorithm is able to capture the investment strategies corresponding to the underlying investor clusters efficiently. One still needs to check that the attribution to experts is working well, i.e. that the investors are mapped to a single, unique expert. To this end, we retrieved from the gating block the attribution probabilities to the $n$ experts of all the investors \textit{a posteriori}. For comparison, we also analyse the investors' embeddings of  Embed-MLP. The comparison of the final embeddings of ExNet-Opt and the ones trained in the Embed-MLP algorithm is shown in Fig. \ref{mlp_vs_exnet}. 

To visualize embeddings, we use here the UMAP algorithm \citep{mcinnes2018umap}, which is particularly relevant as it seeks to preserve the topological structure of the embeddings' data manifold in a lower-dimensional space, thus keeping vectors that are close in the original space close in the embedding space, and making inter-cluster distances meaningful in the two-dimensional plot.
 The two-dimensional map given by UMAP is therefore a helpful tool for understanding how investors relate to each other according to the deep learning method. In these plots, high-activity investors are shown in blue, low-activity investors in red and medium-activity investors in green. We can see in Fig. \ref{mlp_vs_exnet} that the Embed-MLP algorithm did not make a totally clear distinction between the low- and medium-activity clusters, contrarily to the ExNet which separated these two categories with the exception of a few low-activity investors mixed in the medium-activity cluster. The ExNet algorithm was therefore completely able to retrieve the original clusters. 

\begin{figure}[!ht]
	\centerline{\includegraphics[scale = 0.5]{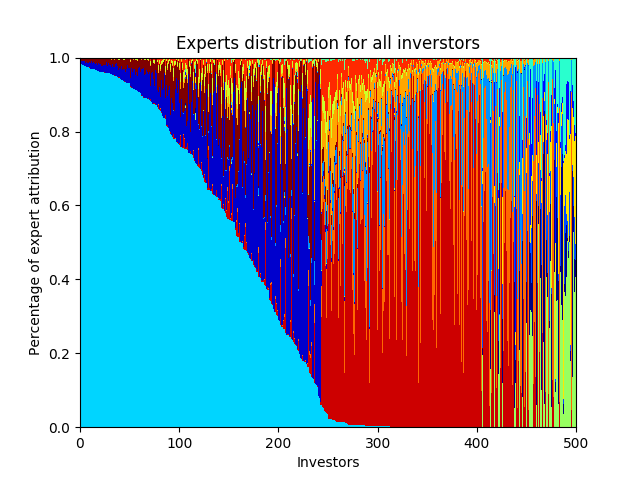}}
	\caption{Distribution over experts of all investors for ExNet-Opt, obtained with random search. Each column shows the attribution probabilities of a given investor, where colors represent experts.}
	\label{exnet_attrib}
\end{figure}

The attribution probabilities to the different experts of ExNet-Opt are shown in Fig. \ref{exnet_attrib}. We see in this figure that the attribution structure of this ExNet instance is quite noisy, with three different behaviours clearly discernable. The first group of investors correspond to the low-activity cluster, the second group to the medium-activity cluster and the last one to the high-activity cluster. Attributions are here very noisy, and investors of a same cluster are not uniquely mapped to an expert.

\begin{figure}[!ht]
	\centerline{\includegraphics[scale = 0.5]{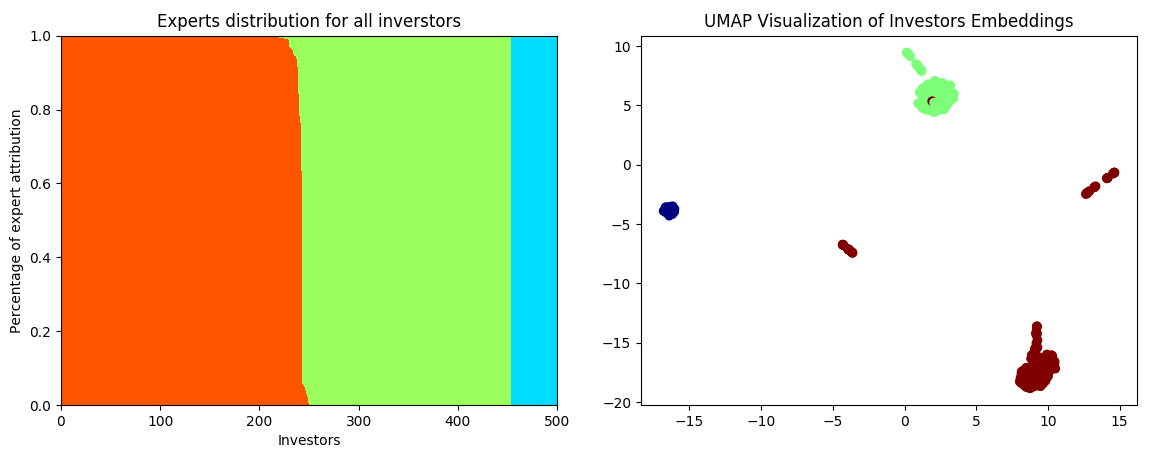}}
	\caption{Distribution over experts of all investors and UMAP visualization of investors embeddings for the ExNet-100 algorithm. Each column of the left plot shows the attribution probabilities of a given investor, where colors represent experts. Colors on the right plot correspond to investors' original clusters: high-activity is shown in blue, medium-activity in green and low-activity in red.}
	\label{exnet100_plots}
\end{figure}

It is however possible to achieve a more satisfactory experts attribution, as one can see in Fig. \ref{exnet100_plots} with the plots of ExNet-$100$. This comes from the fact that the ExNet-$100$ instance used a higher level of $\lambda_{entropy}$ than ExNet-Opt --- at the expense of some performance, we are able to obtain far cleaner attributions to the experts. We see here on the left plot that all investors were attributed almost entirely to one expert only, with for each of the corresponding experts mean attribution probabilities $\overline{p} >99\%$, even with an initial number of experts of $100$, i.e. the $n \gg K$ setting. One can directly see on the UMAP plot three well-defined, monochromatic clusters. We can also see here that a low-activity investor got mixed in the medium-activity cluster, and that two separated low-activity clusters appear --- these separated clusters originate from the fact that some low- and medium-activity investors were marginally attributed to the expert corresponding to the other cluster, as appearing on the experts distribution plot.

The ExNet-$100$ therefore solved the problem that we originally defined, obtaining good predictive performance on the three original clusters and uniquely mapping investors to one expert only, thereby \textit{explicitly} uncovering the initial structure of the investors, a feature that an algorithm such as Embed-MLP is unable to perform.  

\subsection{IBEX data}

\subsubsection{Constructing the dataset}

This experiment uses a real-world,  publicly available dataset published as part of \citet{gutierrez2019mapping} (\url{https://zenodo.org/record/2573031}) which contains data about a few hundred private investors trading 8 Spanish equities from the IBEX index, from January 2000 to October 2007. For a given stock and each day and each investor, the dataset gives the end-of-the-day position,  the open, close, maximum and minimum prices of the stock as well as the traded volume.

We focus here on the stock of the Spanish telecommunication company Telefónica, TEF, as it is the stock with the largest number of trades. Using this data, we try to predict, at each date, whether an investor will be interested into buying TEF or not. An investor is considered to have an interest into buying TEF when $\Delta p^{a}_{t} = p^{a}_{t} - p^{a}_{t-1} > 0$, where $p^{a}_{t}$ is the position of investor $a$ at time $t$. We only consider here the buy interest as the sell interest of private investors can be driven by exogenous factors that cannot be modelled, such as a liquidity shortage of an investor, whereas the buy interest of a investor depends, to some extent, on market conditions. We thus face a binary classification problem which is highly unbalanced: on average, a buy event occurs with a frequency of $2.7\%$.                                                        

We consider a temporal split of our data in three parts: training data is taken from January 2000 to December 2005, validation data from January 2006 to December 2006 and test data from January 2007 to October 2007. We restrict our investor perimeter to investors that bought TEF more than $20$ times during the training period. We build two kinds of features:
\begin{itemize}
\item \textbf{Position features.} Position is shifted such that at date $t$ corresponds $p_{t-1}$, and is normalized for each investor using statistics computed on the training set. This normalized, shifted position is used as is as feature, along with moving averages of it with windows of 1 month, 3 months, 6 months and 1 year.
\item 	\textbf{Technical analysis features.} We compute all the features available in the \texttt{ta} package \citep{talib}, which are grouped under 5 categories: Volume, Volatility, Trend, Momentum and Others features. As most of these features use close price information, we shift them such that features at a date $t$ only use information available up to $t-1$.
\end{itemize}

We are left with $308$ rather active investors and $63$ features.

\subsubsection{Results}
\label{res_ibex}

ExNet and LightGBM are both trained using a combination of random search \citep{bergstra2012random} and hand fine-tuning. Because of the class imbalance of the dataset, the ExNet is trained using the \textit{focal loss} \citep{lin2017focal}, an adaptive re-weighting of the cross-entropy loss. Other popular techniques to handle class imbalance involve undersampling the majority class and/or oversampling the minority one, such as SMOTE \citep{chawla2002smote}. The $\gamma$ parameter of this loss is treated as an hyperparameter of the network, and is also randomly searched. We also used the baseline buy activity rate of each investor in the training period as a benchmark.

\begin{table}[!ht]
\begin{center}
	\begin{tabular}{c c c c}
		\hline
		Algorithm & Train & Val & Test\\ 
		\hline
		\textbf{Historical} & 9.68 & 4.55 & 2.49 \\
		\textbf{LGBM} & 22.22 & 7.53 & 5.35 \\
		\textbf{ExNet-4} & 18.37 & 8.63 & 6.45 \\ \hline
	\end{tabular}
\end{center}
	\caption{Experimental results on IBEX data: average precision scores, expressed in percentage.}
	\label{ibex_results}
\end{table}

The LightGBM shown in Table \ref{ibex_results} used $16$ leaves with a minimum of $128$ samples per leaf, a maximum depth of $4$, a learning rate of $0.002$, a subsample ratio of $35\%$ with a frequence of $5$, a sampling of $85\%$ of columns per tree, with a patience of $50$ for a maximum number of trees of $5000$. The ExNet shown used $4$ experts with two hidden layers of size $32$ and $32$ with a dropout ratio of $50\%$, embeddings of size $d=32$, an input dropout of $10\%$, $\lambda_{spec} = 7.7e^{-4}$
 and $\lambda_{entropy} = 4.2e^{-2}$, a focal loss of parameter $\gamma = 2.5$, a batch size of $1024$, a learning rate of $7.8e^{-4}$ and was trained using Nadam and Lookahead, with an early stopping of patience $20$. As can be seen on this table, both algorithms beat the historical baseline, and the ExNet achieved overall better test performance. While the precision of LightGBM is better in the training set, it is clearly inferior to that of ExNet in the validation set, a sign that ExNet is less prone to overfitting than LightGBM.
 
\begin{figure}[!ht]
	\centerline{\includegraphics[scale = 0.5]{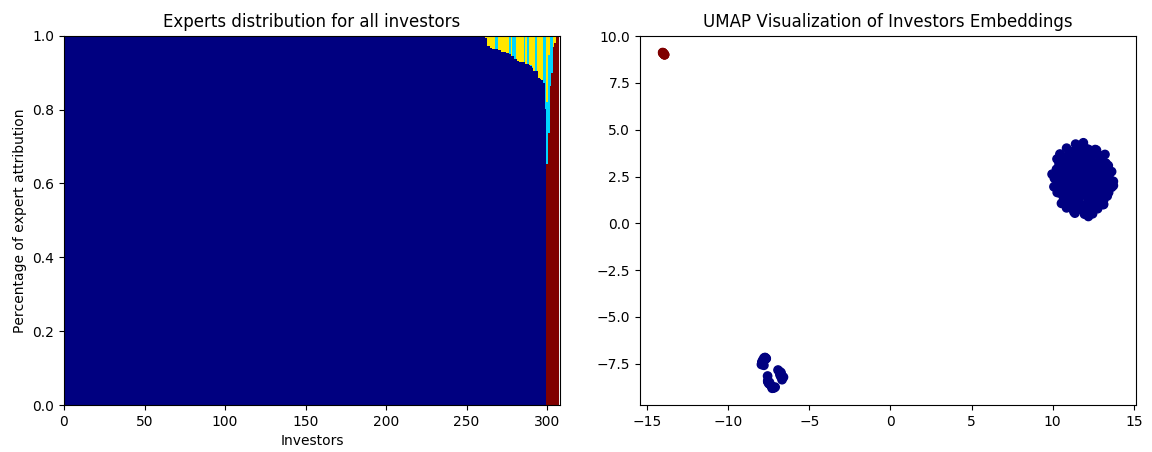}}
	\caption{Distribution over experts of all investors and UMAP visualization of investors embeddings for the ExNet algorithm on the TEF dataset. Each column of the left plot shows the attribution probabilities of a given investor, where colors represent experts --- same colors are used on the right plot.}
	\label{ibex_plots}
\end{figure}

Figure \ref{ibex_plots} gives a deeper view of the results obtained by the ExNet. Three distinct behaviours appear in the left plot. Some of the investors were entirely attributed to the blue expert, some investors used a combination of the blue expert and two others, and some used combinations of the light blue and red experts. These three clusters are remarkably spaced in the UMAP plot on the right. It therefore appears that the ExNet retrieved three distinct behavioural patterns from the investors interacting on the TEF stock, leading to an overall better performance than the LightGBM who was not able to capture them, as the experiments performed in Section \ref{simul_data} show.

\subsubsection{Experts analysis}

We saw in Section \ref{res_ibex} that the ExNet algorithm retrieved three different clusters. Let us investigate in more details what these clusters correspond to.  First, the typical trading frequency of the traders attributed to each of these three clusters are clearly different. The investors that were mainly attributed to the blue expert in Fig. \ref{ibex_plots}, corresponding to the blue cluster on the right of the UMAP plot, can be understood as 'low-activity' investors, trading on average $2.1\%$ of the time. The blue cluster on the left-hand side of the UMAP plot can be understood as medium-activity investors, buying on average $5.5\%$ of the days; the red cluster on the left of the plot is made of high-activity investors ($13.9\%$). The ExNet therefore retrieved particularly well three distinct behaviours, corresponding to three different activity patterns.

To get a better understanding of these three clusters, we can try to assess these clusters' sensitivity to the features used in the model. We use here \textit{permutation importance}, a widespread method in machine learning, whose principle was described for a similar method in \citet{breiman2001random}. The idea is to replace a given feature by permutations of all its values in the inputs, and assess how the performance of the model evolves in that setting. Here, we applied this methodology to the six groups of features: we performed the shuffle a hundred times and averaged the corresponding performance variations. For each of the three clusters, we pick the investor who traded the most frequently, and apply permutation importance to characterize the behaviour of the cluster. Results are reported in Table \ref{ibex_analysis}. 

\begin{table}[!ht]
\begin{center}
	\begin{tabular}{c c c c}
		\hline
		Feature group & Cluster 1 & Cluster 2 & Cluster 3\\ \hline
		\textbf{Position} & \textbf{-37.4\%} & \textbf{-43.1\%} & \textbf{-23.7\%} \\
		\textbf{Volume} & \textbf{-18.6\%} & +10.6\% & \textbf{-19.9\%} \\
		\textbf{Volatility} & \textbf{-22.8\%} & -4.5\% & -2.1\% \\
		\textbf{Trend} & \textbf{-9\%} & -2.4\% & -3.7\% \\
		\textbf{Momentum} & +1.7\% & +4.3\% & \textbf{-13.5\%} \\
		\textbf{Others} & -0.7\% & +8.3\% & +0.2\% \\ \hline
	\end{tabular}
\end{center}
	\caption{Percentage of average precision variation when perturbing features of the group given in the first column for the three clusters appearing in Fig. \ref{ibex_plots}, using permutation importance. Cluster 1 corresponds to the cluster of low-activity investors, cluster 2 to the medium-activity ones and cluster 3 to the high-activity ones.}
	\label{ibex_analysis}
\end{table}

We call in this table cluster 1 the low-activity one, cluster 2 the medium-activity one and cluster 3 the high-activity one. We see that the three groups have different sensibilities to the groups of features that we use in this model. While all clusters are particularly sensitive to position features, the respective sensitivity of groups to the other features vary: leaving aside cluster 2 that only looks sensitive to position, cluster 1 is also sensitive to volume, volatility and trend, whereas cluster 3 is also sensitive to volume and momentum. The clusters therefore not only encode the activity rate, but also the type of information that a strategy needs, and by extension the family of the strategies used by traders, which validate the intuition that underpins the ExNet algorithm.

\subsection{BNPP CIB data}

The previous experiments proved the ability of the network to retrieve the structure of investors with a finite set of fixed investment strategies, and the usefulness of our approach on a real-world dataset. We now give an overview of the results we obtain on the BNPP CIB bonds' RFQ dataset specified in Section \ref{intro} for the non-universality of clients study. 

As a reminder, assets considered are corporate bonds. The data used ranges from early 2017 to the end of 2018 with temporal train/val/test splits, and is made of custom proprietary features using clients-related, assets-related and trades-related data. Our goal is, at a given day, to predict the interest of a given investor into buying and/or selling a given bond; each row of the dataset is therefore indexed by a triplet \textit{(Investor, Bond, Date)}. Targets are constructed as previously explained in Section \ref{intro}. In the experiment conducted here, we consider $1422$ different investors interacting around a total of more than $20000$ distinct bonds. 

\begin{figure}[!ht]
	\centerline{\includegraphics[scale = 0.58]{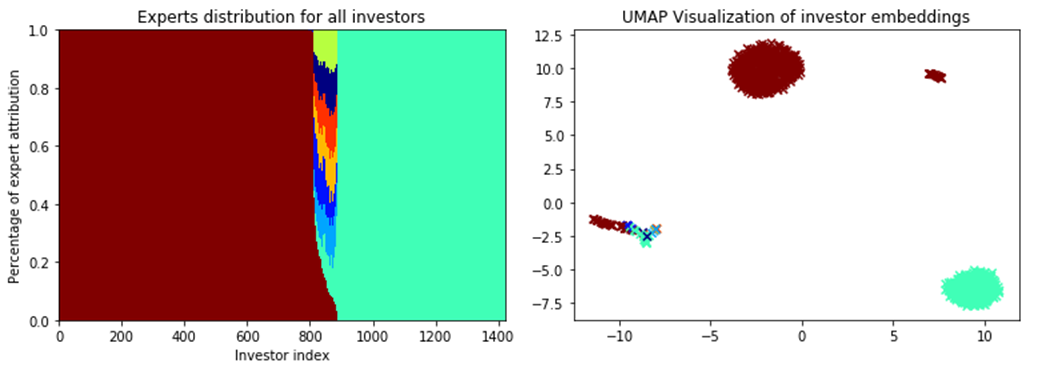}}
	\caption{Distribution over experts of all investors and UMAP visualization of investors embeddings for the ExNet algorithm on the BNPP CIB Bonds' RFQ dataset. Each column of the left plot shows the attribution probabilities of a given investor, where colors represent experts --- same colors are used on the right plot.}
	\label{result_bnpp}
\end{figure}

The left-hand plot of Fig. \ref{result_bnpp} shows the distribution over experts for all the $1422$ considered investors. We see three different patterns appearing : one which used the brown expert only, another one the green expert only and a composite one. These patterns lead to four clusters on the right-hand plot. In this plot, as previously done in the IBEX experiment, each point corresponds to an investor, whose color is determined by the expert to which she is mainly attributed, with colors matching the ones of the left-hand plot. We empirically remark that these clusters all have different activities: the larger brown cluster is twice more active than the green one, the two smaller clusters having in-between average activities. The ExNet therefore retrieved distinct behavioural patterns, confirmed by a global rescaled specialization loss below $0.5$, hence negatively correlated experts.

Obtaining finer clusters could be achieved in multiple ways. A higher-level category could be used as gating input: instead of encoding investors directly, one could encode their sector of activity, in the fashion of the non-universality of clients experiment. With an encoding of the investors, running an ExNet on the investors of one of the retrieved clusters only would also lead to a finer clustering of the investors --- a two-stage gating process could even directly lead to it, and will be part of further investigations on the ExNet algorithm. Note however that these maps (and ExNets) are built from measures of simultaneous distance, hence, do not exploit lead-lag relationships --- how ExNets could be adapted to a temporal setting to retrieve lead-lag relationships will be worthy of future investigations as well.

On a global scale, these plots help us understand how investors relate to each other. Therefore, one can use them to obtain a better understanding of BNP Paribas CIB business, and how BNP Paribas CIB clients' behave on a given market through a thorough analysis of the learnt experts. 

\section{Conclusion}

We introduced a novel algorithm, ExNet, based on the financial intuition that in a given market, investors may act differently when exposed to the same signals, and cluster around a finite number of  investment strategies. This algorithm is able to perform both prediction, be it regression or classification, and clustering at the same time. The fact that these operations are trained simultaneously leads to a clustering that most closely serves the prediction task, and a prediction that is improved by the clustering. Moreover, one can use this clustering \textit{a posteriori}, independently, to gain knowledge as to how individual agents behave and interact with each other. To help the clustering process, we introduced two additional loss terms that penalize the correlation between the inferred investment strategies and the entropy of the investors' allocations to experts. Thanks to an experiment with simulated data, we proved the usefulness of our approach, and we discussed how the ExNet algorithm performs on an open-source dataset of Spanish stock market data and on data from BNP Paribas CIB. Further research on the subject will include how such architectures could be extended and staged, and how they could be adapted to retrieve lead-lag relationships in a given market.

On a final note, the ExNet architecture introduced in this article can be applied  wherever one expects agents to use a finite number of decision patterns, e.g. in  e-shopping or movie opinion databases \citep{bennett2007netflix}.

\section{Acknowledgements}

This work was conducted under the French CIFRE PhD Program, in collaboration between the MICS Laboratory at CentraleSupélec and BNP Paribas CIB Global Markets. We thank Sarah Lemler, Frédéric Abergel and Julien Dinh for helpful discussions and feedback on early drafts of this work. 

\newpage
\bibliographystyle{abbrvnat}
\bibliography{references}

\end{document}